\documentclass[11pt,a4paper]{article}
\usepackage[hyperref]{acl2021}
\usepackage{times}
\usepackage{latexsym}

\usepackage{xcolor}
\usepackage{times}
\usepackage{ragged2e}
\usepackage{amsmath}
\usepackage{bm}
\usepackage{latexsym}
\usepackage{subfigure}
\usepackage{graphicx}
\usepackage{float}
\usepackage{longtable}
\usepackage{booktabs} 
\usepackage{multirow}
\usepackage{amssymb}
\usepackage[ruled,vlined]{algorithm2e}
\usepackage{color, soul}
\usepackage{listings} 

\usepackage{tcolorbox}
\usepackage{cleveref}
\usepackage[hang,flushmargin]{footmisc}
\crefname{section}{§}{§§}
\Crefname{section}{§}{§§}

\definecolor{codegreen}{rgb}{0,0.6,0}
\definecolor{codegray}{rgb}{0.5,0.5,0.5}
\definecolor{codepurple}{rgb}{0.58,0,0.82}
\definecolor{backcolour}{RGB}{252, 253, 246}

\lstdefinestyle{mystyle}{
    backgroundcolor=\color{backcolour},   
    commentstyle=\color{codegreen},
    keywordstyle=\color{magenta},
    numberstyle=\tiny\color{codegray},
    stringstyle=\color{codepurple},
    basicstyle=\ttfamily\footnotesize,
    breakatwhitespace=false,         
    breaklines=true,                 
    captionpos=b,                    
    keepspaces=true,                 
    numbers=left,                    
    numbersep=5pt,                  
    showspaces=false,                
    showstringspaces=false,
    showtabs=false,                  
    tabsize=2
}

\usepackage{microtype}

\aclfinalcopy 

\title{OpenPrompt: An Open-source Framework for Prompt-learning}
\author{Ning Ding\thanks{\quad equal contribution}\hspace{0.5em}, Shengding Hu$^{*}$, Weilin Zhao$^{*}$,  Yulin Chen, \\ \textbf{Zhiyuan Liu}\thanks{\quad corresponding authors}, \textbf{Hai-Tao Zheng}$^{\dag}$, \textbf{Maosong Sun} \\
Tsinghua University, Bejing, China \\
\texttt{\{dingn18, hsd20, zwl19, yl-chen21\}@mails.tsinghua.edu.cn} \\
}

\date{}

\begin{document}
\maketitle
\begin{abstract}
Prompt-learning has become a new paradigm in modern natural language processing, which directly adapts pre-trained language models (PLMs) to $cloze$-style prediction, autoregressive modeling, or sequence to sequence generation, resulting in promising performances on various tasks. 
However, no standard implementation framework of prompt-learning is proposed yet, and most existing prompt-learning codebases, often unregulated, only provide limited implementations for specific scenarios.
Since there are many details such as templating strategy, initializing strategy, and verbalizing strategy, etc. need to be considered in prompt-learning, practitioners face impediments to quickly adapting the desired prompt learning methods to their applications.
In this paper, we present {OpenPrompt}, a unified easy-to-use toolkit to conduct prompt-learning over PLMs. OpenPrompt is a research-friendly framework that is equipped with efficiency, modularity, and extendibility, and its combinability allows the freedom to combine different PLMs, task formats, and prompting modules in a unified paradigm. Users could expediently deploy prompt-learning frameworks and evaluate the generalization of them on different NLP tasks without constraints. OpenPrompt is publicly released at {\url{ https://github.com/thunlp/OpenPrompt}}.




\end{abstract}

\section{Introduction}
\label{sec:intro}

Pre-trained language models (PLMs)~\cite{Han2021PreTrainedMP, qiu2020pre} have been widely proven to be effective in natural language understanding and generation, ushering in a new era of modern natural language processing (NLP). In the early stage of this revolution, a standard approach to adapt PLMs to various specific NLP tasks is the \textit{pretraining-finetuning} paradigm, where additional parameters and task-specific objectives are introduced in the tuning procedure. 
However recently, the paradigm of the adaptation of PLMs is shifting. Originated in T5~\cite{raffel2020exploring} and GPT-3~\cite{brown2020language}, researchers find that PLMs can be effectively stimulated by textual prompts or demonstrations, especially in low-data scenarios. 

Take a simple prompt-based sentiment classification for example, the pipeline consists of a template and a verbalizer, where a template is used to process the original text with some extra tokens, and a verbalizer projects original labels to words in the vocabulary for final prediction. Assume the template is ``\texttt{<text>} \textit{It is} \texttt{<mask>}'', where the token \texttt{<text>} stands for the original text, and the verbalizer is \textit{\{``positive'':``great'', ``negative'':``terrible''\}}.
The sentence ``\textit{Albert Einstein was one of the greatest intellects of his time.}''  will first be wrapped by the pre-defined template as ``\textit{Albert Einstein was one of the greatest intellects of his time. It is} $\texttt{<mask>}$''. The wrapped sentence is then tokenized and fed into a PLM to predict the distribution over vocabulary on the \texttt{<mask>} token position. It is expected that the word \textit{great} should have a larger probability than \textit{terrible}.

As illustrated above, prompt-learning projects the downstream tasks to pre-training objectives for PLMs with the help of textual or soft-encoding prompts. A series of studies of prompt-learning~\cite{Liu2021PretrainPA} have been proposed to investigate the strategies of constructing templates~\cite{schick-schutze-2021-exploiting,gao-etal-2021-making, liu2021gpt}, verbalizers~\cite{hu2021knowledgeable}, optimization~\cite{lester2021power}, and application~\cite{li-liang-2021-prefix, han2021ptr, ding2021prompt} for this paradigm.

\begin{table*}[!th]
\centering
\scalebox{0.85}{
\begin{tabular}{lccccc} 
\toprule 

\textbf{Example}   & \textbf{PLM} & \textbf{Template} & \textbf{Verbalizer}  &  \textbf{Task}  & \textbf{Reference} \\

\midrule

Naive TC & MLM \& Seq2Seq & M. text & M. One-Many & Text Classification & -\\
Naive KP & LM \& Seq2Seq & M. text & - & Knowledge Probing & - \\
Naive FET & MLM & M. text (meta info) & M. One-Many & Entity Typing & \cite{ding2021prompt} \\
PTR & MLM & M. text (complex) & M. One-One & Relation Extratcion & \cite{han2021ptr} \\
P-tuning & LM & Soft tokens & M. One-One & Text Classification & \cite{liu2021gpt} \\
Prefix-tuning & LM, Seq2Seq & Soft tokens & - & Text Generation & \cite{li-liang-2021-prefix}\\
LM-BFF & MLM & A. text & M. One-Many & Text Classification & \cite{gao-etal-2021-making} \\

\bottomrule
\end{tabular}
}
\caption{Some examples implemented by OpenPrompt, where M. is the abbreviation of manually defined and A. is the abbreviation of automatically generated.
Note that different approaches focus on different parts in prompt-learning. Additional to the whole pipeline, our specific implementations of these methods are integrated into the specific classes of OpenPrompt. For example, the core implementation of KPT is in the \texttt{KnowledgeableVerbalizer} class.}
\vspace{-0.3cm}
\label{tab:examples}
\end{table*}



A prompt-learning problem could be regarded as a synthesis of PLMs, human prior knowledge, and specific NLP tasks that need to be handled. Hence, it is hard to support the particular implementations of prompt-learning elegantly with the current deep learning or NLP libraries while there is also a lack of a standard paradigm. 
Previous works pursue the most efficient way to implement prompt-learning with the least modification to the existing framework for traditional fine-tuning, resulting in poor readability and even unstable reproducibility. 
Moreover, the performance of a prompt-learning pipeline varies greatly with the choice of templates and verbalizers~\cite{zhao2021calibrate}, creating more barriers for implementations.
Lastly, there is no comprehensive open-source framework particularly designed for prompt-learning at present, which makes it difficult to try out new methods and make rigorous comparisons for previous approaches.

To this end, we present OpenPrompt, an open-source, easy-to-use, and extensible toolkit for prompt-learning. OpenPrompt modularizes the whole framework of prompt-learning and considers the interactions between each module. We highlight the feature of combinability of OpenPrompt, which supports flexible combinations of diverse task formats, PLMs, and prompting modules. For example, we can easily adapt prefix-tuning~\cite{li-liang-2021-prefix} to a text classification task in OpenPrompt. This feature enables users to assess the generalization of their prompt-learning models on various tasks, but not only the performance on specific tasks.

Specifically, in OpenPrompt, a \texttt{Template} class is used to define or generate textual or soft-encoding templates to wrap the original input. To flexibly support various templates under a unified paradigm, we design a new template language that could easily conduct token-level customization for the corresponding attributes. For example, users can specify which tokens are shared embedding, trainable, or in what way these tokens are to be post-processed, without having to perform complex implementations for specific templates. 
A \texttt{Verbalizer} projects the classification labels to words in the vocabulary, and a \texttt{PromptModel} is responsible for the training and inference process.
Each module in OpenPrompt is clearly defined while retaining its independence and coupling so that researchers can easily deploy a model and make targeted improvements. We also implement baselines with OpenPrompt and evaluate them on a broad scope of NLP tasks, demonstrating the effectiveness of OpenPrompt.

The area of prompt-learning is in the exploratory stage with rapid development. Hopefully, OpenPrompt could help beginners quickly understand prompt-learning, enable researchers to efficiently deploy prompt-learning research pipeline, and empower engineers to readily apply prompt-learning to practical NLP systems to solve real-world problems. OpenPrompt will not only open source all the code, but will also continue to update the documentation to provide detailed tutorials.

\section{Background}

Prompt-learning reveals what the next generation of NLP may look like. 

Although PLMs have achieved tremendous success on almost all the subtasks in NLP, one problem still hangs in the air, \textit{have we really fully exploited the potential of PLMs, especially the big ones?} Conventional fine-tuning uses extra task-specific heads and objectives for adaptation, but this strategy may face two issues.
On the one hand, such an approach creates a natural gap between model tuning and pre-training. On the other hand, as the number of model parameters increases, this fine-tuning approach becomes increasingly difficult to operate due to the massive computational volume (e.g., GPT-3~\cite{brown2020language}). 

By mimicking the process of pre-training, prompt-learning intuitively bridges the gap between pre-training and model tuning. Practically, this paradigm is surprisingly effective in low-data regime~\cite{le2021many, gao-etal-2021-making}. For example, with appropriate template, zero-shot prompt-learning could even outperform 32-shot fine-tuning~\cite{ding2021prompt}. 
Another promising empirical attribute of prompt-learning is the potential to stimulate large-scale PLMs. When it comes to a 10B model, solely optimizing prompts (the parameters of the model are fixed) could achieve comparable performance to full parameter fine-tuning~\cite{lester2021power}. These practical studies imply that we may use prompts to more effectively and efficiently dig the knowledge kept in PLMs, leading to a deeper understanding of the underlying principles of their mechanisms~\cite{wei2021pretrained, qin2021exploring, vu2021spot}.

From a practical implementation point of view, prompt-learning is actually complex and requires a lot of detailed consideration. 
With general-purpose NLP under the prompt-learning paradigm as our target, we present OpenPrompt, a unified toolkit to effectively and efficiently implement prompt-learning approaches.  
OpenPrompt demonstrates a comprehensive view of the programming details of prompt-learning, and enables practitioners to quickly understand the mechanisms and practical attributes of this technique. And one can quickly deploy existing representative prompt-learning algorithms that are already implemented in the package under a unified programming framework. Moreover, OpenPrompt allows researchers or developers to quickly try out new ideas of prompt-learning, which not only includes newly designed templates or verbalizers, but also the exploration of the attributes of prompt-learning, e.g., prompt-based adversarial attacking.

\begin{figure*}[!t]
    \centering
    \includegraphics[width = 0.98\linewidth]{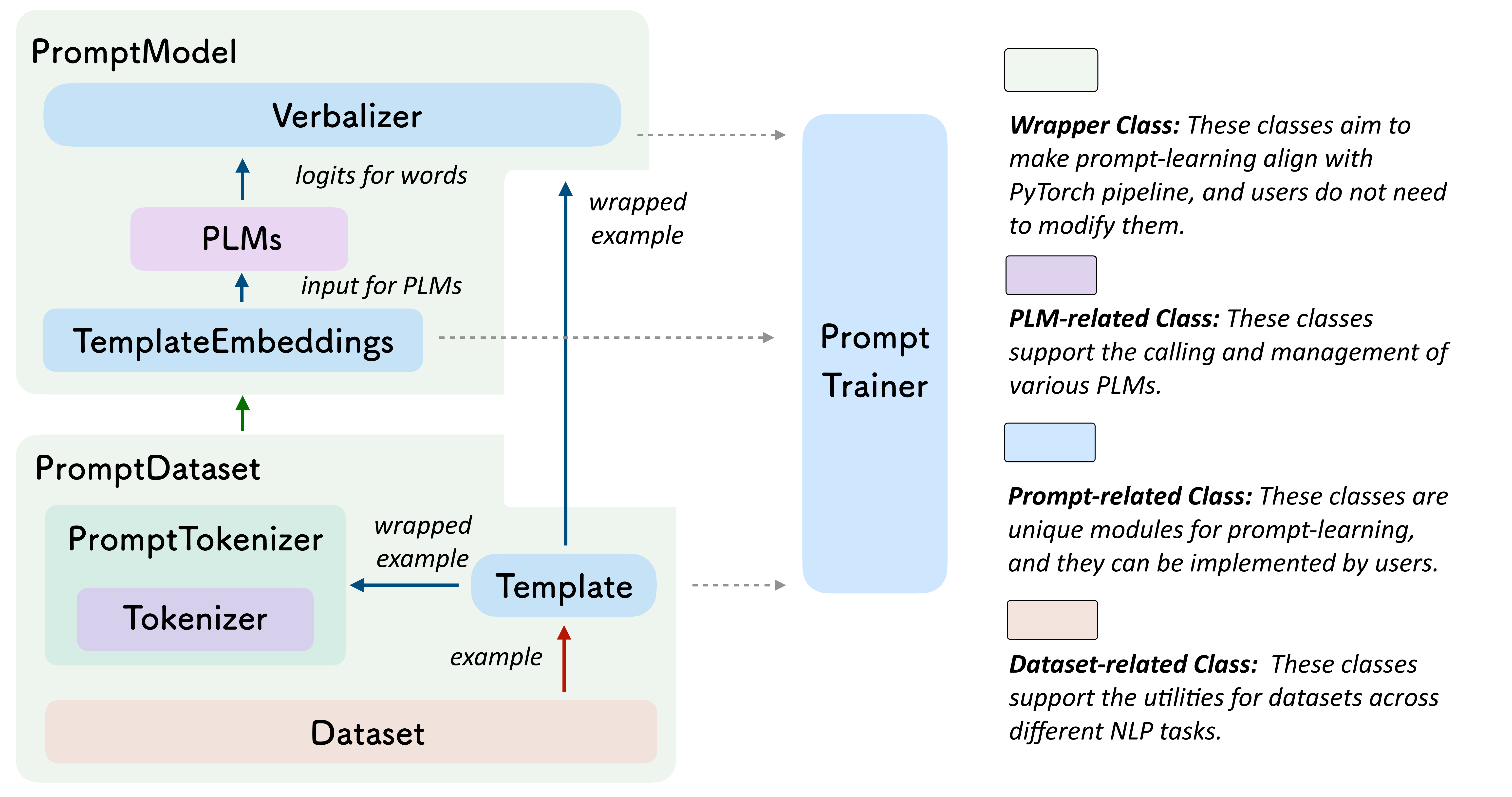}
    \caption{The overall architecture of OpenPrompt. Note that according to the prompt-learning strategies, not all the modules are necessarily used. For example, in generation tasks, there are no verbalizers in the learning procedure. The \texttt{PromptTrainer} is a controller that controls the data flow and the training process with some unique attributes, users can also implement the training process in a conventional fashion.}

    \label{fig:design}
\end{figure*}

\section{Design and Implementation}

As stated in \cref{sec:intro}, prompt-learning is a comprehensive process that combines PLMs, human knowledge, and specific NLP tasks. Keeping that in mind, the design philosophy is to simultaneously consider the independence and mutual coupling of each module. As illustrated in Figure~\ref{fig:design}, OpenPrompt provides the full life-cycle of prompt-learning based on PyTorch~\cite{paszke2019pytorch}. In this section, we first introduce the combinability of OpenPrompt, and then the detailed design and implementation of each component in OpenPrompt.

\subsection{Combinability}

In the NLP world, we usually adopt different PLMs with corresponding objective functions to different underlying tasks (roughly, classification and generation). But in prompt learning, given that the core idea of the framework is to mimic pre-training tasks in the downstream task, which are essentially "predicting words based on context", we can further unify the execution of downstream tasks.
OpenPrompt supports a combination of tasks (classification and generation), PLMs (MLM, LM and Seq2Seq), and prompt modules (different templates and verbalizers) in a flexible way. For example, from a model perspective, T5~\cite{raffel2020exploring} is not only used for span prediction and GPT~\cite{brown2020language} is not only used for generative tasks. From the perspective of prompting, prefix-tuning can also be used for classification, and soft prompt can be used for generation. All these combinations can easily be implemented and validated on NLP tasks in our framework so that we can better understand the mechanisms involved.

\subsection{Pre-trained Language Models}

One core idea of prompt-learning is to use additional context with masked tokens to imitate the pre-training objectives of PLMs and better stimulate these models. Hence, the choice of PLMs is crucial to the whole pipeline of prompt-learning. PLMs could be roughly divided into three groups according to their pre-training objectives. 

The first group of PLMs use masked language modeling (MLM) to reconstruct a sequence corrupted by random masked tokens, where only the losses of the masked tokens are computed. Typical PLMs with MLM objective include BERT~\cite{devlin2018bert}, RoBERTa~\cite{liu2019roberta}, etc, and such an objective is regarded suitable for natural language understanding (NLU). The second group exploits the autoregressive-style language modeling (LM) to predict the current token according to its leading tokens. GPT-3~\cite{brown2020language} is one of the representative works adopting this objective. The third part is the sequence-to-sequence (Seq2Seq) models, which aim to generate a sequence with a decoder conditioned on a separate encoder for an input sequence. Typical seq2seq PLMs include T5~\cite{2020t5}, MASS~\cite{song2019mass} and BART~\cite{lewis2020bart}, etc.

Different PLMs have different attributes, resulting in various adaptation capabilities for different NLP tasks in prompt-learning. 
Practically in OpenPrompt, we support directly loading PLMs from huggingface transformers\footnote{\url{https://huggingface.co/models}}~\cite{wolf2020transformers}, and PLMs implemented by other libraries will be supported in the future. Once the PLM is determined, researchers could deploy a known valid prompt-learning pipeline (e.g., RoBERTa for few-shot sentiment classification) or explore other uses of PLM that could exploit its potential. Users of OpenPrompt do not need to implement objective heads for different PLMs to calculate the corresponding loss, a unified interface can perform these operations automatically (\cref{sec:promptmodel}).

\subsection{Tokenization}

Tokenization is a crucial step in processing data for NLP, and it faces new challenges in prompt-learning. After designing the template, the specific implementation of the tokenization for original input and the designed template could be time-consuming and error-prone. First, in prompt-learning, some specific information such as the indices of entities and masked tokens should be carefully tackled in tokenization. Some small errors, such as the mismatch of masked token indices, may lead to serious consequences. 
Moreover, concatenation and truncation issues after tokenization (templates are not supposed to be truncated) should also be handled. Since different PLMs may have different tokenization strategies, we should also consider the inconsistency in the details of additional context processing.

In OpenPrompt, we specifically design the tokenization module for prompt-learning and significantly simplify the process. By using our encapsulated data processing APIs, users could use the human-readable style to design templates and conveniently operate on the input and the template at the same time. Our component integrates complex information from input and template and then conducts tokenization. Based on the choice of PLMs (MLM, LM, and Seq2Seq), OpenPrompt automatically chooses the appropriate tokenizer in prompt-learning, which could save considerable time for users to process prompt-related data.


\begin{figure*}[!thp]
\centering
\begin{minipage}{0.999\linewidth}
\begin{lstlisting}[language=Python]
# Example A. Hard prompt for topic classification
a {"mask"} news: {"meta": "title"} {"meta": "description"}

# Example B. Hard prompt for entity typing
{"meta": "sentence"}. In this sentence, {"meta": "entity"} is a {"mask"},

# Example C. Soft prompt (initialized by textual tokens)
{"meta": "premise"} {"meta": "hypothesis"} {"soft": "Does the first sentence entails the second ?"} {"mask"} {"soft"}.

# Example D. The power of scale
{"soft": None, "duplicate": 100} {"meta": "text"} {"mask"}

# Example E. Post processing script support 
# e.g. write an lambda expression to strip the final punctuation in data
{"meta": "context", "post_processing": lambda s: s.rstrip(string.punctuation)}. {"soft": "It was"} {"mask"}

# Example F. Mixed prompt with two shared soft tokens
{"meta": "premise"} {"meta": "hypothesis"} {"soft": "Does"} {"soft": "the", "soft_id": 1} first sentence entails {"soft_id": 1} second?

# Example G. Specify the title should not be truncated
a {"mask"} news: {"meta": "title", "shortenable": False} {"meta": "description"}


\end{lstlisting}
\end{minipage} 
\caption{ Some examples of our template language. In our template language, we can use the key ``meta'' to refer the original input text (Example B), parts of the original input (Example A, C, G), or other key information. We can also freely specify which tokens are hard and which are soft (and their initialization strategy). We could assign an id for a soft token to specify which tokens are sharing embeddings (Example F). OpenPrompt also supports the post processing (Example E) for each token, e.g., lambda expression or MLP.
}

\label{fig:code_template} 
\end{figure*}

\subsection{Templates}

As one of the central parts of prompt-learning, a template module wraps the original text with the textual or soft-encoding template. A template normally contains contextual tokens (textual or soft) and masked tokens. In OpenPrompt, all the templates are inherited from a common base class with universal attributes and abstract methods.

Previous works design a wide variety of templates, including manually written template~\cite{schick-schutze-2021-exploiting} and pure soft template~\cite{lester2021power}. ~\citet{gu2021ppt} report a mix of manual template tokens and soft (trainable) tokens sometimes yields better results than separate manual template and soft template. In ~\citet{liu2021gpt}, a promising performance is achieved by fixing the majority of manual tokens while tuning a small number of the others. In ~\citet{han2021ptr}, the template is contextualized, which needs to be filled with the head entity and the tail entity to form a complete one, moreover, the output of multiple positions is used in the loss calculation in their template.  ~\citet{logan2021cutting} design null template with simple concatenation of the inputs and an appended \texttt{<mask>} token. 

It's not reasonable to design a template format for each prompt since it will require high learning cost for practical use. To this end, in OpenPrompt, we design a template language to ease the problem, with which we can construct various types of templates under a unified paradigm.
Our template language takes insight from the dict grammer of Python. And such a design ensures flexibility and clarity at the same time, allowing users to build different prompts with relative ease. 


More specifically, a template node is a text (or empty text) with an attributes' description.  In our template language, one is free to edit the attributes of each token in the template, such as which characters are shared embedding, how the characters are post-processed (e.g. by MLP), etc. We show some template examples in Figure~\ref{fig:code_template}, and the detailed tutorial for writing templates is in our documentation {\url{https://thunlp.github.io/OpenPrompt}}. 

\begin{figure}[!thp]
\centering
\begin{minipage}{0.96\linewidth}
\begin{lstlisting}[language=Python]
from openprompt import ManualVerbalizer

promptVerbalizer = ManualVerbalizer(
    classes = classes,
    label_words = {
        "negative": ["bad"],
        "positive": ["good", "wonderful", "great"],
    },
    tokenizer = bertTokenizer,
)
\end{lstlisting}
\end{minipage} 
\caption{An example to define a Verbalizer, the number of the label words for each class is flexible. }
\vspace{-0.3cm}
\label{fig:code_verbalizer} 
\end{figure}

\begin{figure*}[!ht]
    \centering
    \includegraphics[width = 0.94\linewidth]{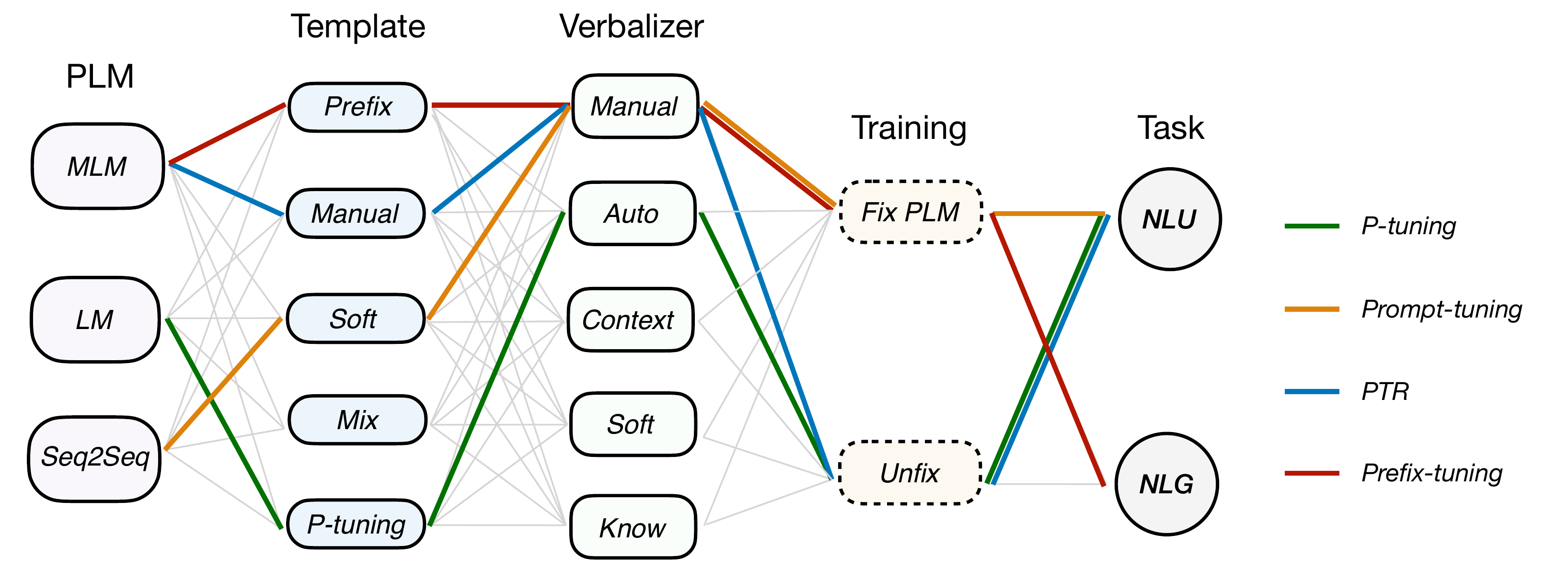}
    \caption{The illustration of the validation space of OpenPrompt. By driving different modules of the framework, we could implement and evaluate different methods on a broad set of NLP tasks. We show four examples in this illustration, the colored lines denote the implementation flow of the corresponding method.}
    \label{fig:space}
    \vspace{-0.3cm}
\end{figure*}

\subsection{Verbalizers}

When it comes to prompt-based classification, a verbalizer class should be constructed to map original labels to label words in the vocabulary.
When a PLM predicts a probability distribution over the vocabulary for one masked position, a verbalizer will extract the logits of label words and integrate the logits of label words to the corresponding class, thereby responsible for the loss calculation. Figure~\ref{fig:code_verbalizer} shows a simple way to define a binary sentiment classification verbalizer.

Similar to templates, all the verbalizer classes are also inherited from a common base class with necessary attributes and abstract methods. Additional to manually-defined verbalizers, we implement automatic verbalizers like AutomaticVerbalizer and KnowledgeableVerbalizer~\cite{hu2021knowledgeable}. Moreover, important operations like calibrations~\cite{zhao2021calibrate} are also realized in OpenPrompt.

\subsection{PromptModel}
\label{sec:promptmodel}

In OpenPrompt, we use a \texttt{PromptModel} object to be responsible for training and inference, which contains a PLM, a \texttt{Template} object, and a \texttt{Verbalizer} object (optional). Users could flexibly combine these modules and define advanced interactions among them. A model-agnostic forward method is implemented in the base class to predict words for the masked positions. One goal of this module is that users do not need to specifically implement heads for different PLMs, but use a unified API to ``predict words for positions that need to be predicted'' regardless of the pre-training objective. An example to define a \texttt{PromptModel} is shown in Figure~\ref{fig:code_promptmodel}.

\begin{figure}[htp]
\centering
\begin{minipage}{0.96\linewidth}
\begin{lstlisting}[language=Python]
from openprompt import PromptForClassification

promptModel = PromptForClassification(
    template = promptTemplate,
    model = bertModel,
    verbalizer = promptVerbalizer,
)

promptModel.eval()
with torch.no_grad():
    for batch in data_loader:
        logits = promptModel(batch)
        preds = torch.argmax(logits, dim = -1)
        print(classes[preds])
\end{lstlisting}
\end{minipage} 
\caption{An example to define a PromptModel and conduct evaluation.}
\vspace{-0.4cm}
\label{fig:code_promptmodel} 
\end{figure}
\subsection{Training}

From the perspective of trainable parameters, the training of prompt-learning could be divided into two types of strategies. The first strategy simultaneously tunes the prompts and the PLM, which is verified to be effective in a low-data regime (OpenPrompt also provides a \texttt{FewshotSampler} to support the few-shot learning scenario).
The second strategy is to only train the parameters of prompts and keep the PLM frozen, this is regarded as a parameter-efficient tuning method and is considered as a promising way to stimulate super-large PLMs.  Both of these strategies can be called with one click in the trainer (or runner) module of OpenPrompt. Trainer modules in OpenPrompt implement training process accompanied with prompt-oriented training tricks, e.g. the ensemble of templates. Meanwhile, OpenPrompt supports experimentation through configuration to easily drive large-scale empirical study.

\section{Evaluation}

OpenPrompt aims to support a broad set of NLP tasks under the paradigm of prompt-learning. In terms of evaluation, we use OpenPrompt to implement various baselines and assess them on the corresponding NLP tasks. We show the validation space in Figure~\ref{fig:space}. And the evaluation tasks include WebNLG~\cite{gardent2017webnlg} for conditional generation, GLUE~\cite{wang2018glue} and SuperGLUE~\cite{wang2019superglue} for natural language understanding; SemEval~\cite{hendrickx2010semeval} for relation extraction; Few-NERD~\cite{ding2021few} for fine-grained entity typing; MNLI~\cite{williams2017broad}, AG's News~\cite{zhang2015character}, DBPedia~\cite{lehmann2015dbpedia} and IMDB~\cite{maas2011learning} for text classification; LAMA~\cite{petroni2019language} for knowledge probing. The processors of these datasets have already been implemented in OpenPrompt, and they are all inherited from a common base \texttt{DataProcessor} class. To keep the results up to date, we are constantly updating and reporting the latest results on our GitHub repository \url{https://github.com/thunlp/OpenPrompt}.

\section{Conclusion and Future Work}

We propose OpenPrompt, a unified, easy-to-use and extensible toolkit for prompt-learning. 
OpenPrompt establishes a unified framework with clearly defined blocks and flexible interactions to support solid research on prompt-learning.
At the application level, OpenPrompt could facilitate researchers and developers to effectively and efficiently deploy prompt-learning pipelines. 
In the future, we will continue to integrate new techniques and features to OpenPrompt to facilitate the research progress of prompt-learning. 

\bibliographystyle{acl_natbib}
\bibliography{acl2021}


\end{document}